\documentclass[sigconf]{acmart}
\usepackage{algorithm}
\usepackage{algorithmic}
\AtBeginDocument{%
  }

\begin{document}

\title{Automatic Deduction Path Learning via Reinforcement Learning with Environmental Correction}

\author{Shuai Xiao}
\authornote{Both authors contributed equally to this research.}
\email{shuai.xsh@alibaba-inc.com}
\affiliation{%
  \institution{Alibaba Group}
  % \streetaddress{P.O. Box 1212}
  \country{China}
}

\author{Chen Pan}
\authornotemark[1]
\email{bopu.pc@antgroup.com}
\affiliation{%
  \institution{Ant Group}
  % \streetaddress{P.O. Box 1212}
  \country{China}
}
% \author{Min Wang}
% \email{fuyue.wm@antgroup.com}
% \affiliation{%
%   \institution{Ant Group}
%   % \streetaddress{P.O. Box 1212}
%   % \city{Dublin}
%   % \state{Ohio}
%   \country{China}
% }
\author{Min Wang, Xinxin Zhu, Siqiao Xue, Jing Wang, Yunhua Hu, James Zhang, Jinghua Feng}
\email{fuyue.wm, qinrui.zxx, siqiao.xsq, lingchen.wj,}
\email{wugou.hyh, james.z, jinghua.fengjh@antgroup.com}
% \author{Xinxin Zhu}
% \email{qinrui.zxx@antgroup.com}
% \author{Siqiao Xue}
% \email{siqiao.xsq@antgroup.com}
% \author{Jing Wang}
% \email{lingchen.wj@antgroup.com}
% \author{Yunhua Hu}
% \email{wugou.hyh@antgroup.com}
% \author{James Zhang}
% \email{james.z@antgroup.com}
% \author{Jinghua Feng}
% \email{jinghua.fengjh@antgroup.com}
\affiliation{%
  \institution{Ant Group}
  % \streetaddress{P.O. Box 1212}
  \country{China}
}

% \affiliation{%

% \affiliation{%

% \affiliation{%

% \affiliation{%

% \affiliation{%

% \affiliation{%

% \affiliation{%

\renewcommand{\shortauthors}{Xiao et al.}

\begin{abstract}
  Automatic bill payment is an important part of business operations in fintech companies. The practice of deduction was mainly based on the total amount or heuristic search by dividing the bill into smaller parts to deduct as much as possible. This article proposes an end-to-end approach of automatically learning the optimal deduction paths (deduction amount in order), which reduces the cost of manual path design and maximizes the amount of  successful deduction. Specifically, in view of the large search space of the paths and the extreme sparsity of historical successful deduction records, we propose a deep hierarchical reinforcement learning approach which abstracts the action into a two-level hierarchical space: an upper agent that determines the number of steps of deductions each day and a lower agent that decides the amount of deduction at each step. In such a way, the action space is structured via prior knowledge and the exploration space is reduced. Moreover, the inherited information incompleteness of the business makes the environment just partially observable. To be precise, the deducted amounts indicate merely the lower bounds of the available account balance. To this end, we formulate the problem as a partially observable Markov decision problem (POMDP) and employ an environment correction algorithm based on the characteristics of the business. In the world's largest electronic payment business, we have verified the effectiveness of this scheme offline and deployed it online to serve millions of users.
\end{abstract}



\keywords{POMDP, 
uncertainty,
hierarchical reinforcement learning,
payment}


\maketitle

\section{Introduction}\label{sec:intro}
With the popularization of Internet finance, more and more people use consumer credit products in various life scenarios \cite{al2020financial, weichert2017future, lu2018decoding}. If the user does not repay the bill in time, there would be a risk of overdue. For a digital payment company, one of the core obligations is to improve profitability by turning monetary losses into benefits and maintain customers' credit scores as well as reduce overdue fee. Therefore, the digital payment company will deduct money from users' authorized account to repay the bill to minimize the negative impacts both on financial companies and customers. 
The traditional institutions usually make full deduction of the bill from the authorized accounts to automatically repay the bill, where the total bill amount is executed~\cite{shoghi2019debt,sanchez2022improving}. This way often fails as the balance is usually lower than the bill amount. Such a restricted strategy hazards the customer's credit report and has overdue fee incurred. An intuitive solution is to divide the bill into several smaller amounts and to deduct them one-by-one sequentially. We call those smaller amounts together with the execution order a {\it deduction path}, and the execution actions {\it intelligent deductions}. In practice, each path has a fixed length limitation considering some technical constraints.

We aim to design an efficient deduction path for each user in a data-driven way, and find out three difficulties. First, as the number of users is huge and the partitions of the bill amounts are numerous, the search space of deduction paths is quite large. In our business, the number of users of credit products usually counts in millions.
Therefore, the underlying search space is huge. Second, the data for learning the path is extremely sparse due to the events of successful deductions are sparse in the historical records. For those customers who need deduction strategies to help them repay bills, the successful deduction records are rare on any consecutive dates. Third, there is an intrinsic property of the problem that the observations of the state of the environment is not complete since the available balance of each account cannot be obtained costless mainly due to the personal data protection protocols.

\begin{figure}[h]
	\centering
	\includegraphics[width=0.8\linewidth]{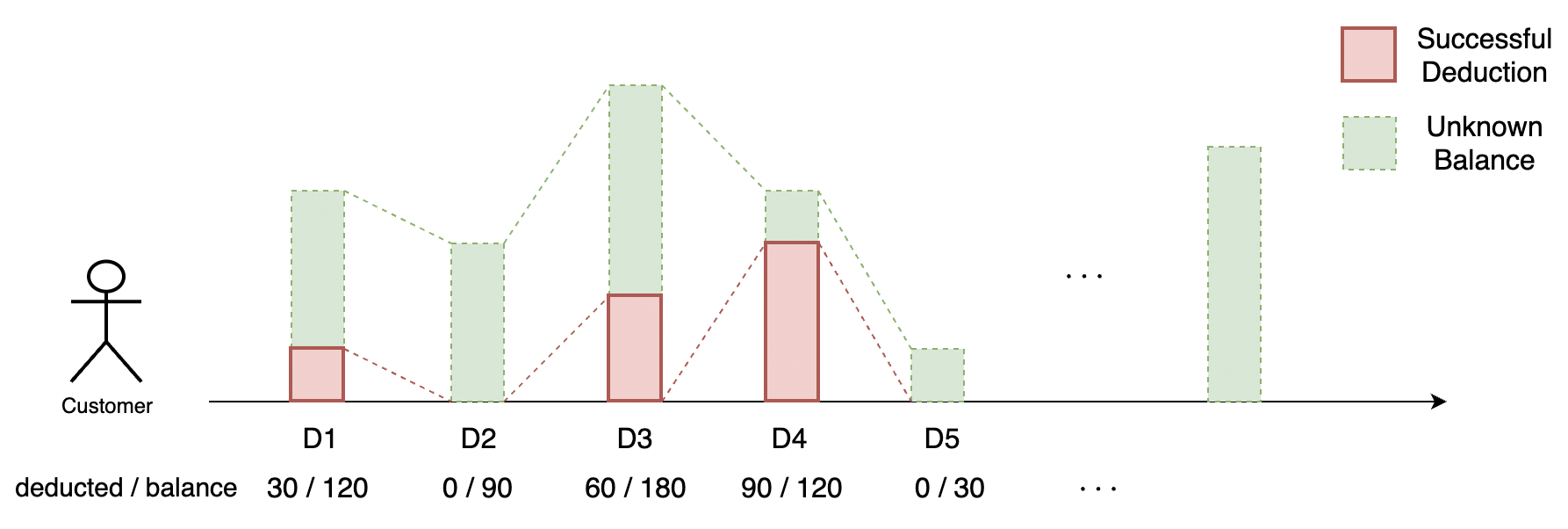}%\vspace{-10pt}
	\caption{An example of partially observed deduction data where observed amount is the lower bound of actual balance.}
	\label{fig:uncertainty}
\end{figure}

In this problem, even with `enough' data, we can only derive a sequence of lower bounds of the dynamically evolved available balance as show in Fig. \ref{fig:uncertainty}, and there is always an unknown gap. 

%Considering the problem as a sequential deduction decision-making problem, we employ reinforcement learning (RL) algorithm to find the optimal path. %We build the RL model based on deep neural networks, which are universal approximators and efficient for RL problems with enormous size of state spaces \cite{NeuroDP}. 
%With the flat RL, it is hard to find a good policy with large search space and rare successfully-deducted feedback. 
To deal with the above problems, we turn to hierarchical RL by a proper abstraction of the deduction task based on the analysis of the deduction business. Firstly, an upper-level agent learns daily sub-goals (how many steps to deduct) and then a lower-level agent decides the deduction amounts series under given sub-goals and current state. On those dates when the balance is too low to be deducted, the number of deduction steps or the amount to deduct can be reduced given the sub-goal outputted by the upper-level agent. The third intrinsic property is caused essentially by incomplete information of the account balance, and we see this as partial observation phenomenon, which is also found in other applications \cite{Murphy2000,hrl_pomdp}. 
We propose a predictive method to correct the observations upwards to approximate the state, so that within a corrected environment a proper algorithm is able to learn a better deduction strategy. Specifically, a structure combining RNN with attention mechanism is built upon the user's payment sequential data as a corrector of the available balance.

There are three contributions of our work:
\begin{enumerate}
    \item First of all, this is the first try in the field of financial technology that deep RL is applied to deduction path learning (for bill payment) as far as we know. This is the an end-to-end automated learning algorithm of the optimal deduction path. The significance of our work is that it provides a well-performed solution for such kind of problems with incomplete information commonly encountered in new businesses of fintech companies.
    
    \item To tackle the difficulty caused by sparsity of successful deduction records, we abstract the problem and propose a hierarchical decision-making approach based on the insight of the business, 
    which helps reduce the invalid decision space and improve the convergence performance of the model.
    
    \item 
    This specific information incomplete problem is characterized as a POMDP, and we reduce the uncertainty via modifying the environment %of RL agent 
    through a correction mechanism. Experiments show that this mechanism is beneficial to increasing the success rate of deduction.
\end{enumerate}

\section{Related Works}
In this part, we introduce related works on intelligent deduction and reinforcement learning under uncertainty.

\subsection{Intelligent Deduction}

Intelligent deduction is an important practice in the modern digital transaction ecosystem. During the debt collection process, 
the institutions can deduct the corresponding amount from the authorized account to repay the loan, which is usually called the exercise of the right of set-off in the industry. Unfortunately, as a third-party payment institution, it cannot obtain the available balance of customers' saving accounts without any cost, then there is a necessity to design a deduction strategy. Related studies mainly focus on predicting potential debtors who are likely to repay. They fall into two categories:  predictive methods~\cite{vasuthevan2021improve, sanchez2022improving} and Markov decision process~\cite{shoghi2019debt,abe2010optimizing}. Predictive models~\cite{vasuthevan2021improve, sanchez2022improving} uses debtors' information and behaviors to predict the possibility of repayment. Markov methods~\cite{shoghi2019debt,abe2010optimizing} also predict debtors' repayment via consecutive estimation, which can capture the the sequential dependencies between actions. The long-term optimization loss also considers the influence of current actions on subsequent states.  Works~\cite{phillips2019artificial,wang2020two} design dialogue agent to interact with debtors to maximize the repayment. Fixed amount full deduction  is initiated based on the user's bill~\cite{shoghi2019debt,sanchez2022improving}. Manually-designed heuristic search~\cite{tarjan1972depth,chin2006improving}
for deduction paths often requires expert knowledge and suffers from the model misspecification problem~\cite{uppal2003model, hansen2006robust,bonhomme2018minimizing} which means the path developed does not match the real data distribution and is not efficient.

\subsection{Reinforcement Learning}
Reinforcement learning is naturally employed to solve sequential decision problems \cite{RL:Intro}. 

RL has been proven successful in many applications~\cite{silver2018general,mnih2013playing,xue_meta_2022}, such as Go, Atari games, protein structure prediction. In finance, the application of RL is not very extensive because training an RL model relies on a large amount of data, and it is costly to collect enough data. Existing applications of RL in finance include asset pricing, portfolio selection, risk measurement, trading strategies~\cite{RL_finance22,survey_rlfin21}, where RL acts as optimal control agent.
Widely-used RL algorithms include actor-critic~\cite{konda2000actor} and  DQN~\cite{van2016deep,mnih2015human} for processing discrete actions.

Hierarchical reinforcement learning (HRL)~\cite{pateria2021hierarchical} is introduced to handle the challenge of exploration that results from the large state and action spaces, which is one of the difficulties we face.
Especially, a hierarchical DQN (h-DQN) model is proposed  to solve the RL problems with vary sparse and delayed rewards \cite{HRL_DQN16}.

Partially observable Markov decision process (POMDP) is often chosen to model the real world problem that the agent can only obtain incomplete information of the environment.
Since the true state is not obtainable, a hidden variable called belief state is introduced as a substitution for the agent to learn policies. The belief state depends on the historical observations and actions, then an RNN structured model is ideal to characterize it, which helps maintain a long-term memory of the historical information \cite{Murphy2000, DTQN22}.

DQNs are usually employed to learn optimal policies in POMDPs. Egorov used a DQN to map belief states to an optimal action in \cite{DRL_pomdp15}. There are also various ways of combining RNNs with DQNs in POMDPs, which mainly address the shortcomings of the agent's limited memory of historical state. The authors in \cite{DRQN15} investigated the effects of adding recurrency to a DQN via an LSTM, and shown advantages in the scenario that partial state information is missing in some video games. A deep attention recurrent Q-network (DARQN) is built in \cite{DARQN15} that adds an attention mechanism to leverage the LSTM's representation of the historical information. Most recently, a deep transformer Q-network (DTQN) is introduced to dealing with POMDPs in \cite{DTQN22}, which replaces the recurrent layers with a transformer decoder structure with position encodings.

\section{Problem Definition}

As one of the world's largest financial technology companies, its consumer credit products have a great number of users. After the monthly bill is issued, the user should complete the repayment no later than the due date.
In order to reduce the negative impact of overdue, the institution will initiate a deduction to the user's authorized saving account(s) after the due date to help the user repay the bill if she/he forgets to repay it in time. With this service, the company could improve the users' experience and help them maintain good credit scores in the nation's central bank system.

The implementation of the deduction tasks relies on a batch processing system to send requests to different bank card issuers simultaneously with a cost $c$ at each step.
Therefore, the decision problem is set to maximize the total amount of successful deductions within a limited steps, so as to reduce the negative impacts on users.

\begin{figure}[t]
	\centering
 \includegraphics[width=0.8\linewidth]{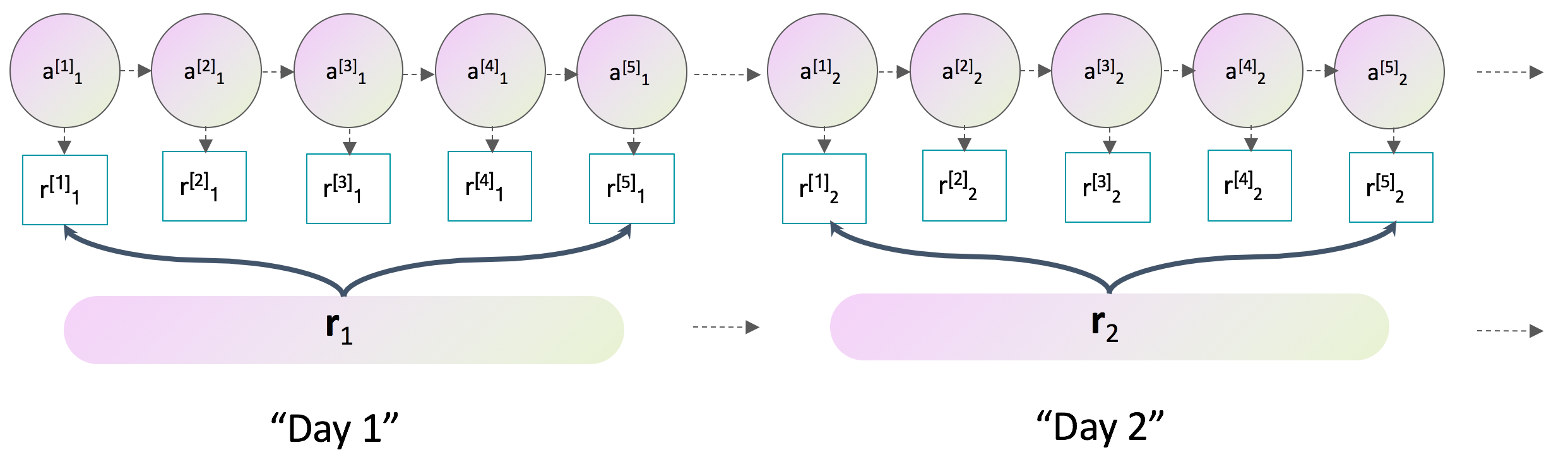}%\vspace{-10pt}
	\caption{The deduction procedure on consecutive dates. 
 }
	\label{fig:hrl}
\end{figure}

In the execution practice, multiple deduction steps (usually with given step limitation, say, five steps) would be made to each account every day. As shown in Fig.~\ref{fig:hrl}, the amount is $a_t^i$ for the $i$-th deduction on day $t$. Let the available balance in the account be denoted by $y_{real}$, if $y_{real}\ge a_t^i$, then the deduction succeeds, and the deduction amount $r_t^i=a_t^i$. After such deduction, the balance becomes $y_{real}-a_t^i$. It would fail, if $y_{real}< a_t^i$. This is the reason that the observed deductions are the lower bounds of the available balance. 

The objective of intelligent deduction is to maximize the total deduction amount, which is formulated as follows
\begin{small}
\begin{equation}\label{equ:total}
     \max_{\{N(t),r_t^{i}\} \sim \pi}{y} := \sum_{t=1}^T \sum_{i=0}^{N(t)}  (r_t^{i}-c),
\end{equation}    
\end{small}
where $r_t^{i}$ is deducted amount at step $i$ on day $t$,  $\pi$ is the deduction path model, and $N(t)$ is number of deduction steps on day $t$.

\begin{figure}[h]
	\centering
	\includegraphics[width=0.8\linewidth]{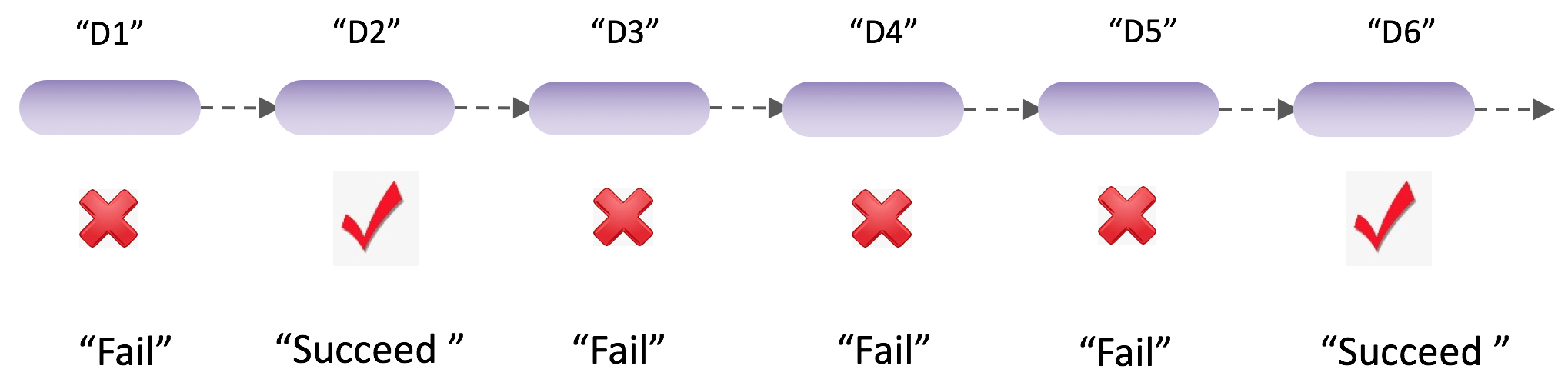}%\vspace{-10pt}
	\caption{An example of the sparsity of successful deductions.}
	\label{fig:path}
\end{figure}

The difficulty of the problem does not only include the uncertainty of the underlying balance as mentioned above, but also the sparsity of the successful deduction historical data. In other words, the feedback obtained from the environment is extremely sparse. In the experimental part, we will give a detailed analysis why the positive feedback is so sparse. In real scenarios, we found that the percentage of successful deductions every day is very small (lower than 30\%). As shown in Fig.~\ref{fig:path}, in normal circumstances, deductions for an account often fail on multiple consecutive dates.

\section{Deduction Path Learning}
After studying the characteristics of the deduction business, we find two outstanding issues which limit the potential performance of data-driven deduction learning. First, the historical deduction data which are used for learning the deduction path are very sparse. Deductions with a high percentage of users fail on consecutive days, which renders the poor convergence of local minima for deduction path algorithm.
Second, we also notice the fact that the deducted amount merely indicates a lower bound of the available balance. We collect thousands of available balance of account through collaboration with banks and find the our previous deduction strategy only deduct 80\% of the available balance on average. 
For this phenomenon, we design two strategies to 
improve the performance. First, we take advantage of user's historical behavior data to predict the available balance of account which can serve as a proxy of available balance for training the path learning agent.

Second, the action spaces are abstracted in two-layered hierarchical structure.

\begin{figure}[h]
	\centering
	\includegraphics[width=0.9\linewidth]{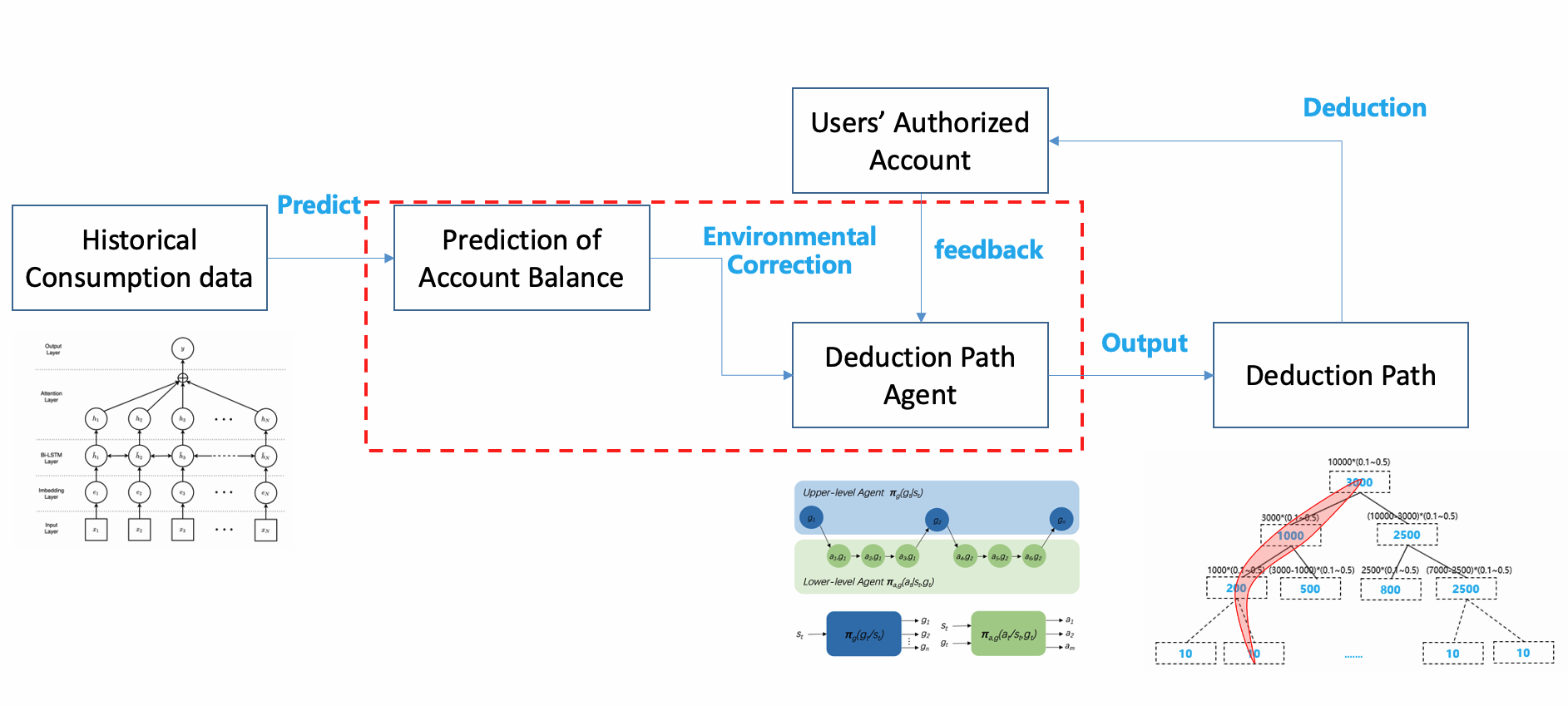}%\vspace{-10pt}
	\caption{The learning and decision process of automatic deduction path learning.}
	\label{fig:sys}
\end{figure}

The overall learning and decision process of deduction path can be illustrated in the paradigm of Fig.~\ref{fig:sys}. The environment is corrected through an RNN estimator and the deduction path agent generates deduction path and updates its parameters after receiving feedback from corrected environments. After convergence, the agent outputs optimal deduction path. 

The learning algorithm is summarized in Algo.~\ref{alg:main_method}.

\begin{algorithm}[tb]
%\small
    \caption{Learning algorithm for intelligent deduction path 
    % finding via environmental correction and action abstraction
    }\label{alg:main_method}
    
    %\REQUIRE
    % \STATE \textbf{Input}
    \textbf{Input}:
    Historical consumption sequence data and successfully deducted amounts\\
    \textbf{Parameter}: Initialization for upper-level agent $\mathrm{Q_1}$ and lower-level agent $\mathrm{Q_2}$
    \begin{algorithmic}[1] 
    \STATE Initialize replay buffer $\mathrm{D}_1$ for $\mathrm{Q_1}$ and $\mathrm{D}_2$ for $\mathrm{Q_2}$;
    %\ENSURE
        \WHILE{Not converge }
            \STATE Update environmental correction using Eq.~\eqref{equ:amt_pred};
        \ENDWHILE
        \WHILE{Not converge }
             \STATE For historical deduction trajectories, using corrected deducted amounts via Eq.~\eqref{equ:env-corrected}.
             \STATE Select a subtask $g$ via upper-level agent  $\mathrm{Q_1}$ with exploration.
             \STATE For each deduction steps in one day, iteratively select action $a$ using lower-level agent $\mathrm{Q_2}$ and collect interior reward from corrected environment. Store experience tuple $(s,g),a^i,r_t^i,(s^{\prime},g)$ in buffer $\mathrm{D}_2$.
             \STATE  Store experience tuple $(s,g,r_t, s^{\prime})$ in buffer $\mathrm{D}_1$. 
             \STATE Update agents $\mathrm{Q_1}$ and $\mathrm{Q_2}$ using the loss with Eq.'s~\eqref{equ:q1} and~\eqref{equ:q2}.
        \ENDWHILE
  
    \end{algorithmic}
    \normalsize
\end{algorithm}

\subsection{Environment Correction: Balance Prediction}

In order to obtain a relatively accurate environmental feedback, we built an environmental correction model.
Data analysis shows that there is a high positive correlation between users' historical consumption data and the available balance in the account. Therefore, we utilize the Bi-LSTM model~\cite{650093} with the attention mechanism to predict the available balance based on the consumption data. 

\begin{figure}[h]
	\centering
	\includegraphics[width=0.8\linewidth]{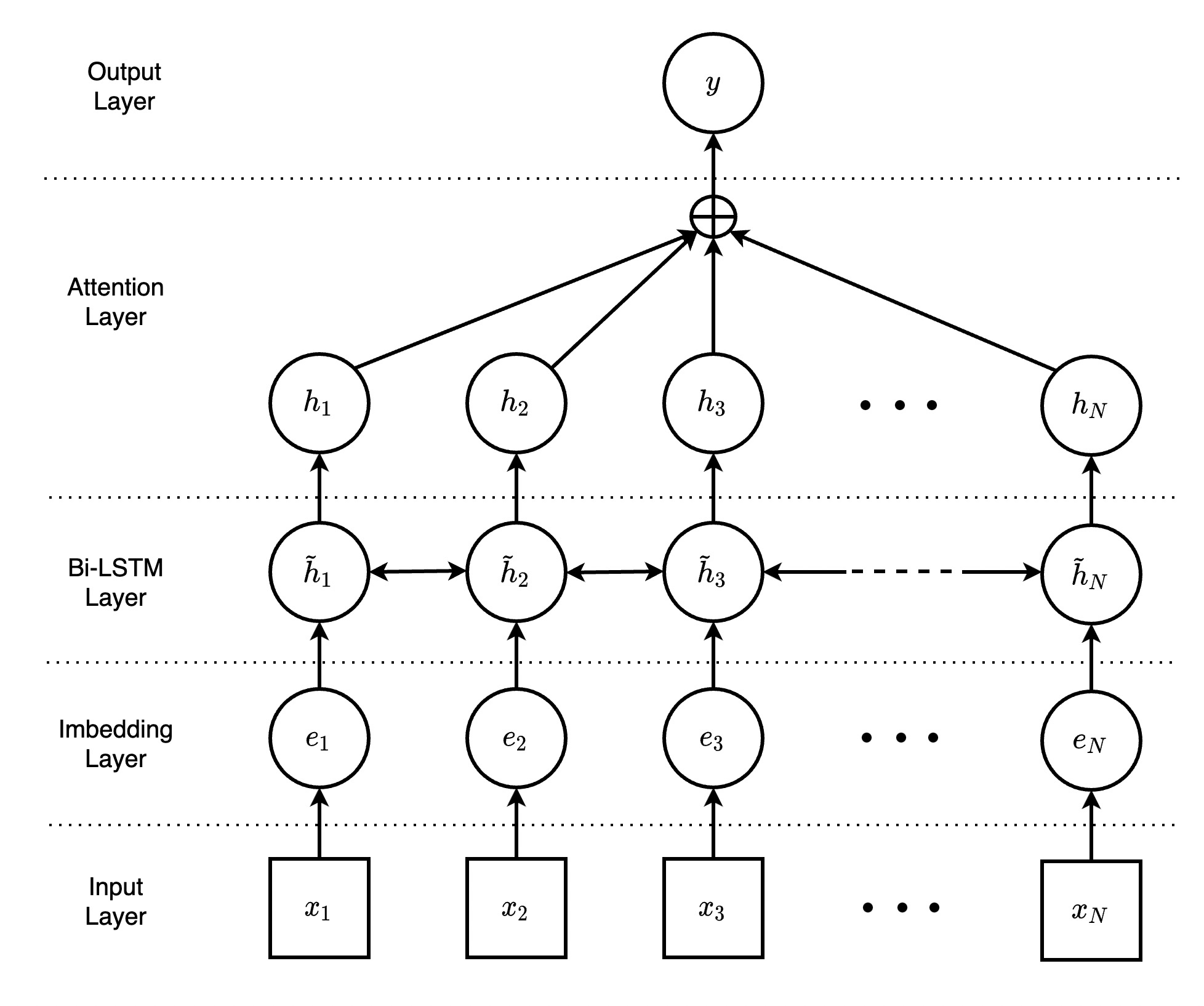}%\vspace{-10pt}
	\caption{The network structure of deduction prediction.}
	\label{fig:bi-lstm}
\end{figure}

Specifically, we extract the user's consumption behavior data before deduction as the input of the model to predict the available balance. The historical consumption data is embedded and transformed using Bi-LSTM as shown in Fig.~\ref{fig:bi-lstm}. Then the final output is calculated by weighted sum of historical consumption amounts. The weight coefficient of each payment is computed through the attention mechanism.
The specific implementation is as follows
\begin{equation}
     y_{pred} = \sum_{i=1}^N w_i  x_i,
     \label{equ:amt_pred}
\end{equation}
where $\{x_i,i=1,\dots,N\}$ is the sequence of historical consumption data and $w_i$ is the attention weight of the $i$-th value. $y_{pred}$ is the estimated target for real deduction amount. $x_i$ together with other features including transaction types $\kappa$ (consumption or payment) and timestamps $t$ are firstly embedded and then transformed through Bi-LSTM networks to derive $h_i$ as illustrated in Fig.~\ref{fig:bi-lstm}. Therefore, the embeddings of inputs can be written as $e_i = \text{Imbed}([x_i, \kappa_i, t_i ])$. Then the hidden vector $\tilde{h}=[\overrightarrow{h}, \overleftarrow{h}]$ of Bi-LSTM can be computed through $\overrightarrow{h}_{i+1} = LSTM(\overrightarrow{h}_{i}, e_i)$ and $\overleftarrow{h}_{i} = LSTM(\overleftarrow{h}_{i+1}, e_{i+1})$. The weights $w_i$ are calculated using attention mechanism
\begin{equation}
     w_{i} = \frac{e^{W h_i +b}}{\sum_{j=0}^N e^{W h_j +b}},
\end{equation}
where $W, b$ are learnable parameters.

We use the model to learn the relationship between the available balance on the deduction day and the consumption behavior before the deduction day. As mentioned before, if the deduction path learning model directly uses the amount of deduction as the goal, it will lead to the problem of underestimating the deduction amount. Therefore, consumption data of the saving account is brought in as the potential incremental space for deduction. 

In practice, since the successfully deducted amount $y_{deducted}$  is a lower bound of the available balance $y_{real}$, the predicted balance $y_{pred}$ is used as a correction term. Thus $y_{deducted}$  and $y_{pred}$ are added together as an approximation of $y_{real}$. We do hope that the correction term (i.e., $y_{pred}$) added in this way is within a reasonable range, and ideally should account for a small proportion of the available balance. If the correction item is too small, it will not work as it should do, and if it is too large, it may be distorted. The approximation of the available balance,  $y_{corrected}$, is formulated as
\begin{equation}
    y_{corrected} = y_{deducted} + \alpha\cdot y_{pred},
    \label{equ:env-corrected}
\end{equation}
where the parameter $\alpha$ is used to adjust the weight of the correction term. This hyperparameter will be selected via an ablation test, and the details will provided in the following experiment section.

With this approximation $y_{corrected}$ in hand, while the model in the second part is learning the deduction path, $y_{corrected}$ is treated as a substitution of the available balance. During the  deduction process, if the deduction amount is less than $y_{corrected}$, we regard this deduction as successful, and the remaining balance is $y_{corrected}$ minus the amount deducted in the current step.

\subsection{Deduction Path Learning: Action Abstraction}

Here, we employ a data-driven approach with proper action abstraction to automatically generate each deduction path on the corrected environment derived from the first part as described in the former section. The method is self-adaptive with little human intervention during deployment in the operation and fit for the sparse positive feedback characteristic of deduction business.

Since the successful deduction events are too sparse, directly applying flat RL to such problem is problematic. First, the positive feedback sparsity causes the value network converge slowly. Second, the rewards span in a long delayed horizon, and this makes training samples in a short horizon unbalanced. To reduce those impact, we abstract deduction action into a two-layered hierarchical space: the upper-level action determines the deduction times $N(t), 0,\ldots,5$ on day $t$ and the lower-level action decides deduction amount $a_t^i$ at each step. The explored space can be reduced significantly in a large time-scale space via this proper abstraction. An upper-agent first learns a sub-goal for each day, and then determines the sequential actions in the next day based on the proposed sub-goals and feedback from the lower-agent. In such way, the deduction path agent is able to abstract the deduction knowledge at multiple levels. Moreover, the action space can be cut off based on the sub-goal. To be concrete, an upper-level agent proposes a subtask to complete and a lower-level agent chooses a primitive action under current state given the goal. The subtask can be defined by specific purposes, such as the number of deduction steps on one day.

\begin{figure}[h]
	\centering
	\includegraphics[width=0.8\linewidth]{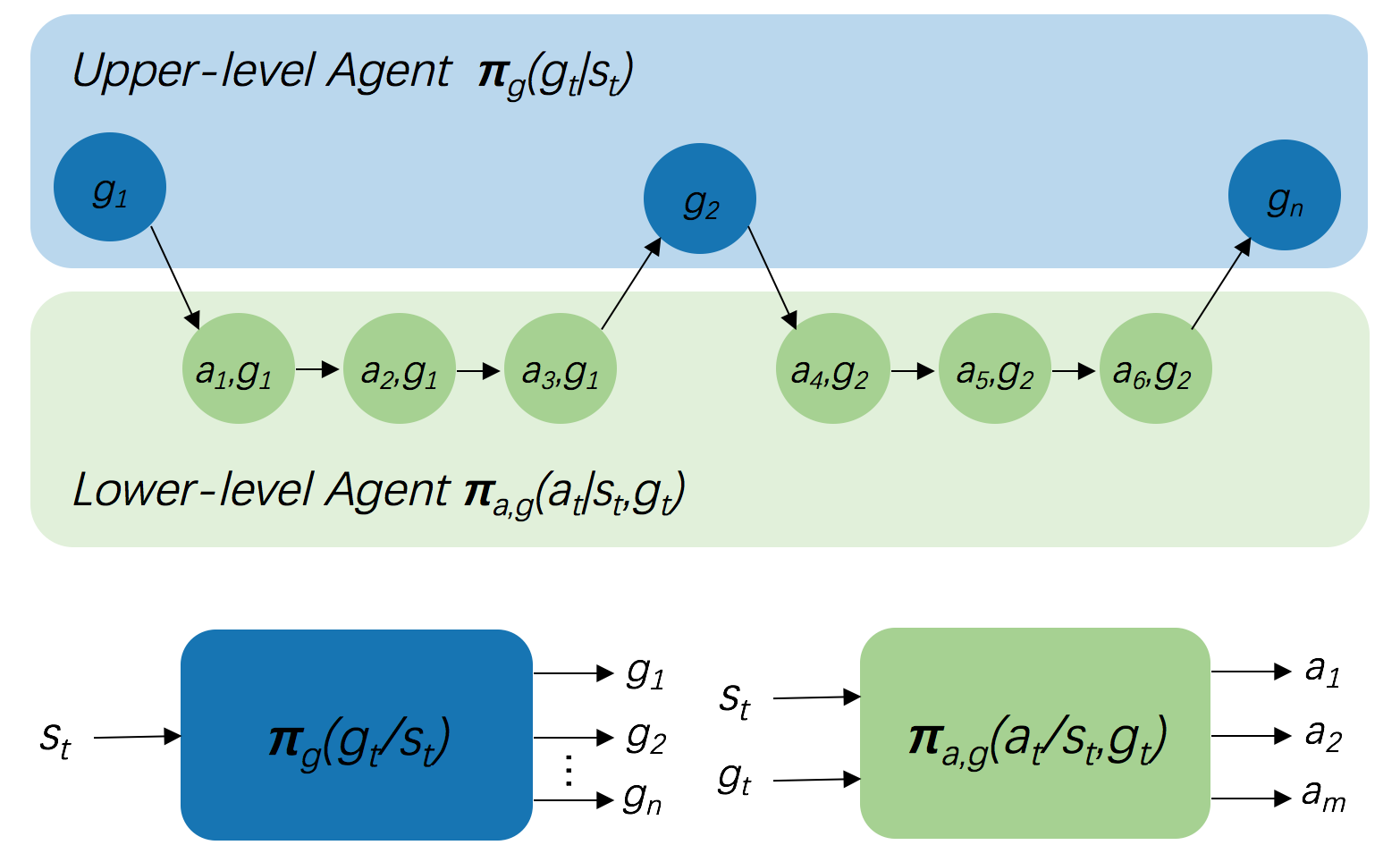}%\vspace{-10pt}
	\caption{The decision process of multi-level action abstraction for deduction task. 
 % The upper-level agent proposes a subtask (i.e, 0 or 3 steps) to complete and the lower-level agent chooses a primitive action (deduction amount at each step) under current state given the goal. 
 }
	\label{fig:two}
\end{figure}
As shown in Fig.~\ref{fig:two}, the upper-level agent 
%$\mathrm{Q}_1$ 
learns whether it could successfully deduct the money some day as a subtask $g$ and output deduct times $N(t)$. 
The upper-level agent is parameterized by $\theta_1$ with an LSTM network which embeds previous actions and results and three-layer feed-forward NNs. 
With given state $s$ of the environment, the policy $\pi_g (g|s)$ of choosing subtasks' results from the value function of upper-level agent
 $\mathrm{Q}_1$ with parameters $\theta_1$. Under the guidance of this subtask $g$, the lower-level agent
$\mathrm{Q}_2$ with parameters $\theta_2$
 adjusts the deduction amount $a$ each step. Similarly, the policy $\pi_{a,g}(a|s^\prime,g)$ for determining the deduction amount results from the value function $\mathrm{Q}_2$ of lower-level agent. $\mathrm{Q}_2(s^\prime,a|g,\theta_2)$ is the value function provided the deduction amount equals $a$ at state $s^\prime$. The lower agent is with an LSTM network which embeds previous actions and results and three-layer feed-forward NNs. Those value functions are provided as follows from the learning objective of Eq.~\eqref{equ:total},
 the value functions of upper-level and lower-level agents are formulated as Bellman's equations
\begin{small}
\begin{equation}
\label{equ:q1}
   \mathrm{Q}_1 (s,g) 
    = \underset{\pi_{\theta_1}}{\max}\mathbf{E}\left[ \sum_{i=0}^{N(t)}(r_t^i -c)
  + \eta \max_{g^{\prime}} \mathrm{Q}_1 (s_{t}, g^{\prime})\bigg | \begin{matrix} s_t = s,\\ g_t =g,\end{matrix} \pi_{\theta_1} \right],
\end{equation}
\begin{equation}
\label{equ:q2}
     \mathrm{Q}_2 (s^\prime,a|g)
     = \underset{\pi_{\theta_2}}{\max} \mathbf{E}\left[ r_{t}^i -c
      + \gamma\max_{a_{t}^{i+1}} \mathrm{Q}_2 (s^\prime_{t+1}, a_{t}^{i+1}| g) \Bigg| \begin{matrix} s^\prime_t = s^\prime, \\ a_{t}^{i}=a,\\ g_t =g, \end{matrix} \,\pi_{\theta_2} \right] ,
\end{equation}
\end{small}
% \begin{equation}
% \label{equ:q1}
% \begin{aligned}
%    & \mathrm{Q}_1 (s,g) \\
%     = &\underset{\pi_{\theta_1}}{\max}\mathbf{E}\left[ \sum_{i=0}^{N(t)}(r_t^i -c)
%   + \eta \max_{g^{\prime}} \mathrm{Q}_1 (s_{t}, g^{\prime})\bigg | \begin{matrix} s_t = s,\\ g_t =g,\end{matrix} \pi_{\theta_1} \right],
% \end{aligned}
% \end{equation}
% \begin{equation}
% \label{equ:q2}
% \begin{aligned}
%      &\mathrm{Q}_2 (s^\prime,a|g)\\
%      = &\underset{\pi_{\theta_2}}{\max} \mathbf{E}\left[ r_{t}^i -c
%       + \gamma\max_{a_{t}^{i+1}} \mathrm{Q}_2 (s^\prime_{t+1}, a_{t}^{i+1}| g) \Bigg| \begin{matrix} s^\prime_t = s^\prime, \\ a_{t}^{i}=a,\\ g_t =g, \end{matrix} \,\pi_{\theta_2} \right] ,
% \end{aligned}
% \end{equation}
where $r_t^i$ is the reward (i.e., the deducted amount) at $i$-th step on day $t$ and $c$ is the constant cost at each step, and two scalars $\eta,\gamma\in (0,1]$ are the discounting factors.

The upper-level agent learns whether the account can be successfully deducted on a certain day and output the proper deduction times $N(t)$, e.g., 0 for failed days and 5 for potentially successful days, and the feedback from the environment (i.e., reward $r_t^i$) is the sum of the deduction amount. The lower-level agent determines the deduction amount for each step based on the judgment of the upper-level agent. The feedback (i.e., reward $r_t^i$) of the lower-level agent is given by the corrected environment and is equal to $a_t^i -c $ if $a_t^i\le y_{corrected}$ or $-c$ if the deduction fails.

The state for upper-level agent is described as  $s= (U_p, U_a, t, H_t)$, 
where $U_p$ represents user profile features, such as gender, age, cities and income, etc. $U_a$ means user's activities, such as payment frequency, transfer frequency, etc.,
$t$ represents the day when the agent performs deductions. $H_t$ is the records of historical deduction including actions and results.
The state for lower agent is described as  $s^\prime=$ ($U_p$, $U_a$, \textit{$i$}, $H_t$, $g$, $B$), where $i$ is the $i$-th step of deduction, $B$ is the remaining debt balance which equals the bill amount minus the total amount that has been deducted, and $g$ is the action of upper-level agent. For upper-level agent, its action space has six sub-tasks $g$, i.e., $(0,1,\dots, 5)$ times, where 0 means no deduction would be done on that day. For each lower-level agent, its action space contains $50$ actions, and the deduction amount corresponding to the action $a_t^i$ when it succeeds is
$r_t^i = \frac{a_t^i}{50}\times B$.
% \begin{equation}
%     r_t^i = \frac{a_t^i}{50}\times B.
% \end{equation}
The reward is $r_t^i-c$ when it succeeds otherwise $-c$.

\section{Experiment and Deployment}

In this section, the offline/online results of practical deduction methods are provided and the deployment of proposed method in real productive setting is discussed.

\subsection{Experimental Settings}
\subsubsection{Baselines}
We evaluated the performance of models against several practical baselines.

\begin{itemize}
    \item \textbf{Full deduction} Full deduction is a common operation in practice where the the total amount of the bill is deducted once~\cite{shoghi2019debt,sanchez2022improving}.
    \item \textbf{Heuristic search}  The widely-used  heuristic search practice~\cite{tarjan1972depth,chin2006improving} is designed like this: A fixed number of deduction steps is pre-defined at each turn. At the first step, half of the bill amount would be executed. If it fails, at the next step half of the amount of the first step will be executed; if it succeeds, half of the amount of the remaining bill amount will be executed. And at the subsequent steps, similar execution logic would be conducted until it reaches the limited steps. 
    %This heuristic method is actually not optimal.
    \item \textbf{Predictive deduction} Similar to predictive methods~\cite{vasuthevan2021improve, sanchez2022improving}, we use a four-layered DNN (with hidden sizes 1024, 512, 256)  to predict the balance of bank account and deduct the predicted amount. 
    \item \textbf{DQN} Like works~\cite{shoghi2019debt,abe2010optimizing}.  The DQN uses a LSTM network to
embed previous actions and results,  and then concatenate with other features. Then a three-layer
feed-forward NNs (with hidden sizes 512, 256)  are used as the value prediction network.
    \item \textbf{Variants of the proposed method}
    Ablation tests against degraded versions of the proposed model DQN with action abstraction and corrected environment (DQN-A2CE) including DQN with corrected environment (DQN-CE), DQN with action abstraction (DQN-A2) are also conducted. 
The DQN and DQN-CE share the same architecture with a LSTM network which embeds previous actions and results and three-layer feed-forward NNs (with hidden sizes 512, 256). The DQN-A2CE and DQN-A2 share the same network architecture whose upper and lower agents have a LSTM network which embeds previous actions and results followed by three-layer feed-forward NNs (with hidden sizes 512, 256).
    
\end{itemize}

\subsubsection{Evaluation Metric}
 We define the `success rate' of deduction as the ratio between the total amount deducted $y_{deducted}$ minus the total cost $C$ and the total bill amount $y_{bill}$: $\text{SuccRate} = {(y_{deducted}-C)}/{y_{bill}}$.
 % \begin{small}
 % \begin{equation}
 %     \text{SuccRate} = \frac{y_{deducted}-C}{y_{bill}}.
 % \end{equation}
 % \end{small}
 % where $y_{deducted}$ is the amount deducted while $C$ is the total cost and $y_{bill}$ is the available balance in the account.

\subsection{The Prediction of Account Balance}

Because of the hardness of obtaining the available balance of the customers' bank accounts, and the sparsity of historical successful deduction records.
%, it would result in underestimated deduction paths if the algorithm relies on the observed data. 
We, therefore, collected thousands sample accounts with available balance through a collaboration with some banks, which is costly and private. The statistics of the collected data are shown in Table~\ref{tab:fea_impor} which are divided into training and evaluation datasets for the prediction.

\begin{table}[!tb]
\centering
\caption{Statistic of training and evaluation datasets collected from banks for available balance prediction.}
\resizebox{0.45\textwidth}{!}{
\begin{tabular}{rr}
%   \addlinespace
    \hline
    Training set & 134,032\\
     Evaluation set& 64,124\\
     Average amount deducted from the account&  71 units\\
     Average of account consumption amount& 500 units\\
     Average of account consumption times& 3 times\\
    \hline
%   \bottomrule
\end{tabular}}
\label{tab:fea_impor}
\end{table}

The prediction accuracy is measured by MAPE (Mean Absolute Percentage Error) ( $\text{Avg}_i((y^i_{pred} - y^i_{real})/y^i_{real})$).  The MAPE of the attention based model is 0.12, which is low enough when considering the noise and difficulty of predicting account balance. 

\begin{figure}[h]
	\centering
	\includegraphics[width=0.8\linewidth]{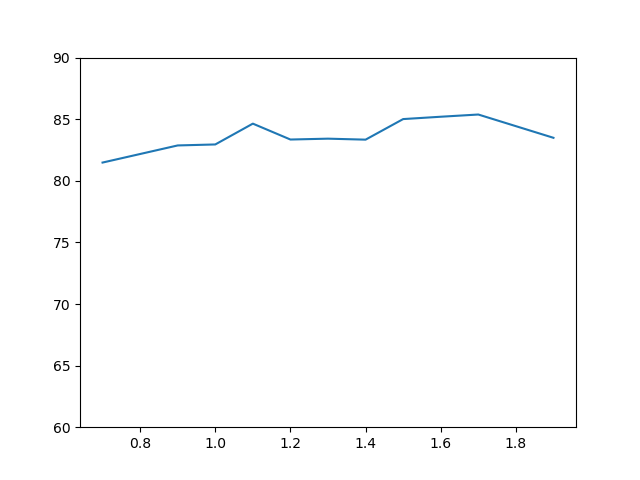}%\vspace{-10pt}
	\caption{The ablation test of hyper-parameter $\alpha$.
 }
	\label{fig:coef}
\end{figure}

The coefficient $\alpha$ in Eq.~\eqref{equ:env-corrected} as a hyperparameter is searched with the deduction successful rate as an indicator. The searching results are shown in Fig.~\ref{fig:coef}. The horizontal axis is the values of $\alpha$, and the vertical axis is the deduction success rate. One can see that the model performs well when this parameter is greater than $1.1$. We choose $\alpha=1.6$ here.

\subsection{Results and Discussions}
Data analysis shows
a serious imbalance between account balance and the amount of pending deductions. Most executions of deduction fail in history as the amount of pending deductions is much larger than the account balance.

The results of the deduction path learned via different algorithms are shown in Table~\ref{tab:succ_rate}. One can see that the automatically learned deduction path via DQN is better than the expert-designed heuristic method and DNN. The success rate of deduction is significantly improved. The model decides the deduction amount to be executed next time based on the previous deduction information. In this way, the algorithm can search for multiple times to approximate the available balance of the user's account and maximize the total amount of deduction. This method needs not rely on expert experience to design the deduction strategy, which avoids the problem of model misspecification, and it is also able to adaptively learn the optimal path from the data.
With the proper action abstraction which reduces the exploration space and difficulty of learning, DQN-A2 improves the success rate from 78.9\% to 82.1\%.
The success rate of DQN-A2CE rises from 82.1\% to 85.4\%, which is the contribution of the correction mechanism to the environment.

\begin{table}[h]
\centering
\caption{The offline results of the deduction path learned via different algorithms.}
% \resizebox{0.3\textwidth}{!}
\begin{small}
{
\begin{tabular}{r|c c c c}
%   \addlinespace
    \bottomrule
    Model	& Full Deduction & Heuristic Search & DNN & DQN \\
    \hline
    SuccRate & 68.1\% (0.000) & 76.4\% (0.000) &   74.2\% (0.035)  &	78.9\% (0.043) \\
    \toprule
    \bottomrule
    Model	& DQN-CE & DQN-A2	& DQN-A2CE \\
    \hline
    SuccRate & 79.6\% (0.041)	& 82.1\% (0.052) &	85.4\% (0.049)\\
  \toprule
\end{tabular}}
\end{small}
\label{tab:succ_rate}
\end{table}

For the model deployment and maintenance, the environment correction model used to predict the available balance is updated once a month. Due to the frequent change of the balance, the deduction path learning is trained and updated once a week. In the inference phase, the deduction strategy of DQN-A2CE is provided offline one day ahead. During online A/B test for one month, the online comparisons are shown in Table~\ref{tab:succ_rate_online}. Successfully deducted amount of DQN-A2CE raises 9\% compared to expert-designed binary search.  From Nov. 2021 till present, DQN-A2CE has been deployed in production environment and generates deduction paths for millions of users over more than one year, which gains significantly economic profit and reduce the negative impact on users.

\begin{table}[h]
\centering
\caption{The offline results of the deduction path learned via different algorithms.}
% \resizebox{0.3\textwidth}{!}
\begin{small}
{
\begin{tabular}{r|c c c c}
%   \addlinespace
    \bottomrule
    Model	& Full Deduction & Heuristic Search & DNN & DQN \\
    \hline
    SuccRate & 70.6\%  & 77.8\%  &  75.1\%   &	79.4\%  \\
    \toprule
    \bottomrule
    Model	& DQN-CE & DQN-A2	& DQN-A2CE \\
    \hline
    SuccRate & 80.5\%	& 83.0\%  &	85.7\% \\
  \toprule
\end{tabular}}
\end{small}
\label{tab:succ_rate_online}
\end{table}  
\section{Conclusion}

To conclude, our model increases the success rate of deductions significantly comparing to the manual designation approach as well as those vanilla RL algorithms. This method shows its efficiency even when the available balance is of specific uncertainty (i.e., the historical deduction amounts are only the lower bounds of the available balance), and the  successful deduction records are extremely sparse. As a further work, a problem worthy studying is to build a big model to estimate the available balance to efficiently represent other behaviors of customers and variables of the economic environment. Another interesting direction is to apply this automatic path searching method to other financial technology scenarios to reduce labor costs and improve production efficiency, say in the field of logistics.

\clearpage
% References
\bibliographystyle{ACM-Reference-Format}
\bibliography{cikm23}
\end{document}